\pdfoutput=1

\documentclass[11pt]{article}

\usepackage{naacl2021}

\usepackage{times}
\usepackage{latexsym}
\usepackage{graphicx}
\usepackage{hyperref}
\usepackage{array}
\usepackage{xcolor}
\usepackage{color, colortbl}
\usepackage[T1]{fontenc}

\usepackage[utf8]{inputenc}

\usepackage{microtype}

\title{Identifying Automatically Generated Headlines using Transformers}

\author{Antonis Maronikolakis \\
  CIS, LMU Munich \\
  \footnotesize{\texttt{antmarakis@cis.lmu.de}} \\\And
  Hinrich Schutze \\
  CIS, LMU Munich  \\\And
  Mark Stevenson \\
  University of Sheffield \\
  \footnotesize{\texttt{mark.stevenson@sheffield.ac.uk}} \\}

\begin{document}
\maketitle
\begin{abstract}
False information spread via the internet and social media influences public opinion and user activity, while generative models enable fake content to be generated faster and more cheaply than had previously been possible. In the not so distant future, identifying fake content generated by deep learning models will play a key role in protecting users from misinformation. To this end, a dataset containing human and computer-generated headlines was created and a user study indicated that humans were only able to identify the fake headlines in $47.8\%$ of the cases. However, the most accurate automatic approach, transformers, achieved an overall accuracy of $85.7\%$, indicating that content generated from language models can be filtered out accurately.
\end{abstract}

\section{Introduction}
\label{sec:intro}

Fake content has been rapidly spreading across the internet and social media, misinforming and affecting users' opinion \cite{survey,survey_acm}. Such content includes fake news articles\footnote{For example, \href{https://www.bbc.com/news/election-us-2020-54838684}{How a misleading post went from the fringes to Trump's Twitter}.} and truth obfuscation campaigns\footnote{For example, \href{https://www.latimes.com/world-nation/story/2019-12-16/taiwan-the-new-frontier-of-disinformation-battles-chinese-fake-news-as-elections-approach}{Can fact-checkers save Taiwan from a flood of Chinese fake news?}}. While much of this content is being written by paid writers \cite{fake_it}, content generated by automated systems is rising. Models can produce text on a far greater scale than it is possible to manually, with a corresponding increase in the potential to influence public opinion. There is therefore a need for methods that can distinguish between human and computer-generated text, to filter out deceiving content before it reaches a wider audience.

While text generation models have received consistent attention from the public as well as from the academic community \cite{plugnplay, adversarial_gen}, interest in the detection of automatically generated text has only arisen more recently \cite{automatic_detection_survey}. Generative models have several shortcomings and their output text has characteristics that distinguish it from human-written text, including lower variance and smaller vocabulary (\citet{text_degen_orig, gtlr}). These differences between real and generated text can be used by pattern recognition models to differentiate between the two. In this paper we test this hypothesis by training classifiers to detect headlines generated by a pretrained GPT-2 model \cite{gpt2}. Headlines were chosen as it has been shown that shorter generated text is harder to identify than longer content \cite{humans_fooled}.

The work described in this paper is split into two parts: the creation of a dataset containing headlines written by both humans and machines and training of classifiers to distinguish between them. The dataset is created using real headlines from the \textit{Million Headlines} corpus\footnote{\href{https://www.kaggle.com/therohk/million-headlines}{Accessed 25/01/2021.}} and headlines generated by a pretrained GPT-2. The training and development sets consist of headlines from 2015 while the testing set consists of 2016 and 2017 headlines. A series of baselines and deep learning models were tested. Neural methods were found to outperform humans, with transformers being almost $35\%$ more accurate.

Our research highlights how difficult it is for humans to identify computer-generated content, but that the problem can ultimately be tackled using automated approaches. This suggests that automatic methods for content analysis could play an important role in supporting readers to understand the veracity of content. The main contributions of this work are the development of a novel fake content identification task based on news headlines\footnote{Code available at \href{http://bit.ly/ant_headlines}{http://bit.ly/ant\_headlines}.} and analysis of human evaluation and machine learning approaches to the problem.

\section{Relevant Work}

\citet{survey} compiled a survey on fake content on the internet, providing an overview of how false information targets users and how automatic detection models operate. The sharing of false information is boosted by the natural susceptibility of humans to believe such information. \citet{automatic_detection} and \citet{opinion_spam} reported  that humans are able to identify fake content with an accuracy between 50\% and 75\%. Information that is well presented, using long text with limited errors, was shown to deceive the majority of readers. The ability of humans to detect machine-generated text was evaluated by \citet{real_or_fake}, showing that humans struggle at the task.

\citet{text_degen_orig} investigated the pitfalls of automatic text generation, showing that sampling methods such as Beam search can lead to  low quality and repetitive text. \citet{gtlr} showed that automatic text generation models use a more limited vocabulary than humans, tending to avoid low-probability words more often. Consequently, text written by humans tends to exhibit more variation than that generated by models. 

In \citet{grover}, neural fake news detection and generation are jointly examined in an adversarial setting. Their model, called Grover, achieves an accuracy of 92$\%$ when identifying real from generated news articles. Human evaluation though is lacking, so the potential of Grover to fool human readers has not been thoroughly explored. In \citet{gpt3}, news articles generated by their largest model (175B parameters) managed to fool humans 48\% of the time. The model, though, is prohibitively large to be applied at scale. Further, \citet{humans_fooled} showed that shorter text is harder to detect, both for humans and machines. So even though news headlines are a very potent weapon in the hands of fake news spreaders, it has not been yet examined how difficult it is for humans and models to detect machine-generated headlines.

\section{Dataset}\label{sec:data}

\subsection{Dataset Development} 

The dataset was created using Australian Broadcasting Corporation headlines and headlines generated from a model. A pretrained\footnote{As found \href{https://huggingface.co/gpt2}{in the HuggingFace library.}} GPT-2 model \cite{gpt2} was finetuned on the headlines data. Text was generated using sampling with temperature and continuously re-feeding words into the model until the end token is generated.

Data was split in two sets, 2015 and 2016/2017, denoting the sets a ``defender" and an ``attacker" would use. The goal of the attacker is to fool readers, whereas the defender wants to filter out the generated headlines of the attacker. Headlines were generated separately for each set and then merged with the corresponding real headlines.

The ``defender" set contains $72,401$ real and $414,373$ generated headlines, while the ``attacker" set contains $179,880$ real and $517,932$ generated.

\subsection{Dataset Analysis}

Comparison of the real and automatically generated headlines revealed broad similarities between the distribution of lexical terms, sentence length and POS tag distribution, as shown below. This indicates that the language models are indeed able to capture patterns in the original data. 

\begin{figure}[ht]
  \includegraphics[width=0.45\textwidth]{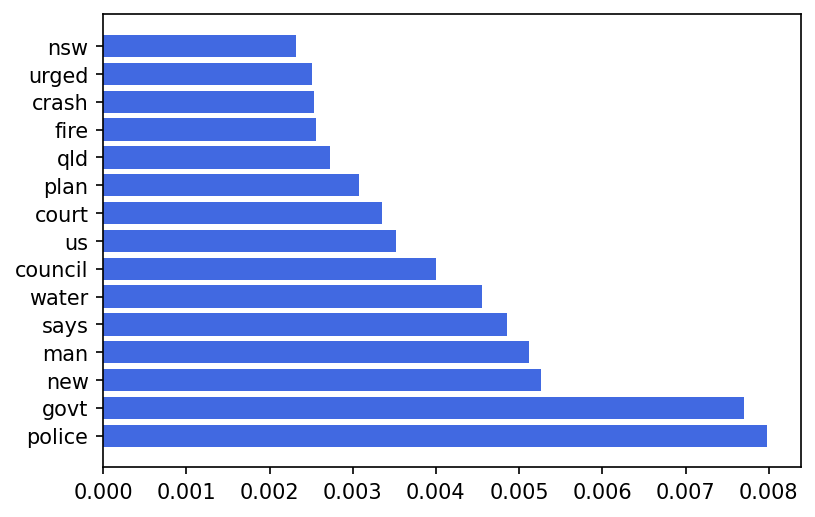}
  \caption{Top 15 Words for real headlines}
  \label{top_real}
\end{figure}

\begin{figure}[ht]
  \includegraphics[width=0.45\textwidth]{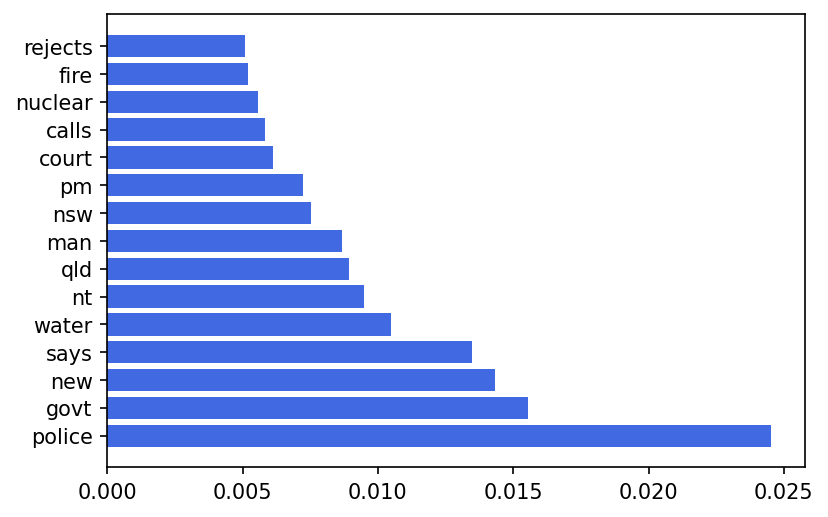}
  \caption{Top 15 Words for generated headlines}
  \label{top_fake}
\end{figure}

Even though the number of words in the generated headlines is bound by the maximum number of words learned in the corresponding language model, the distribution of words is similar across real and generated headlines. In Figures \ref{top_real} and \ref{top_fake} we indicatively show the 15 most frequent words in the real and generated headlines respectively. POS tag frequencies are shown in Table \ref{table_ling} for the top tags in each set. In real headlines, nouns are used more often, whereas in generated headlines the distribution is smoother, consistent with findings in \citet{gtlr}. Furthermore, in generated headlines verbs appear more often in their base (VB) and third-person singular (VBZ) form while in real headlines verb tags are more uniformly distributed. Overall, GPT-2 has accurately learned the real distribution, with similarities across the board.

Lastly, the real headlines are shorter than the generated ones, with $6.9$ and $7.2$ words respectively.

\begin{table}[!t]
\small
\centering
\begin{tabular}{|l|l|l|l|}
\hline
\rowcolor[HTML]{a8a8a8} 
\multicolumn{2}{|c|}{\cellcolor[HTML]{a8a8a8}Real}                                                     & \multicolumn{2}{c|}{\cellcolor[HTML]{a8a8a8}Generated}                                                   \\ \hline
\rowcolor[HTML]{b5b5b5} 
\multicolumn{1}{|c|}{\cellcolor[HTML]{b5b5b5}POS} & \multicolumn{1}{c|}{\cellcolor[HTML]{b5b5b5}freq} & \multicolumn{1}{c|}{\cellcolor[HTML]{b5b5b5}POS} & \multicolumn{1}{c|}{\cellcolor[HTML]{b5b5b5}freq}                                                                                                                                                     \\ \hline
NN & 0.372 & NN & 0.352 \\ \hline
NNS & 0.129 & NNS & 0.115 \\ \hline
JJ & 0.109 & JJ & 0.113 \\ \hline
IN & 0.108 & IN & 0.113 \\ \hline
VB & 0.045 & VB & 0.061 \\ \hline
TO & 0.040 & TO & 0.056 \\ \hline
VBZ & 0.033 & VBZ & 0.047 \\ \hline
VBP & 0.031 & VBP & 0.022 \\ \hline
VBN & 0.020 & RB & 0.017 \\ \hline
VBG & 0.020 & VBG & 0.015 \\ \hline

\end{tabular}
\caption{Frequencies for the top 10 part-of-speech tags in real and generated headlines}
\label{table_ling}
\end{table}

\subsection{Survey}

A crowd-sourced survey\footnote{Participants were students and staff members in a mailing list from the University of Sheffield.} was conducted to determine how realistic the generated text is. Participants (n=$124$) were presented with $93$ headlines (three sets of 31) in a random order and asked to judge whether they were real or generated. The headlines were chosen at random from the ``attacker" (2016/2017) headlines.

In total, there were 3435 answers to the `real or generated' questions and 1731 (50.4\%) were correct. When presented with a computer-generated headline, participants answered correctly in 1113 out of 2329 (47.8\%) times. In total 45 generated headlines were presented and out of those, 23 were identified as computer-generated (based on average response). This is an indication that GPT-2 can indeed generate realistic-looking headlines that fool readers. When presented with actual headlines, participants answered correctly in 618 out of 1106 times (55.9\%). In total 30 real headlines were presented and out of those, 20 were correctly identified as real (based on average response).

Of the 45 generated headlines, five were marked as real by over 80\% of the participants, while for the real headlines, 2 out of 30 reached that threshold. The five generated headlines were:

\begin{center}
	\textsf{\footnotesize{Rumsfeld Talks Up Anti Terrorism Campaign}}\\
	\textsf{\footnotesize{Cooper Rebounds From Olympic Disappointment}}\\
	\textsf{\footnotesize{Jennifer Aniston Tops Celebrity Power Poll}}\\
	\textsf{\footnotesize{Extra Surveillance Announced For WA Coast}} \\
	\textsf{\footnotesize{Police Crack Down On Driving Offences}}
\end{center}

At the other end of the spectrum, there were seven generated headlines that over 80\% of the participants correctly identified as being computer-generated:

\begin{center}
	\textsf{\footnotesize{Violence Restricting Rescue Of Australian}}\\
	\textsf{\footnotesize{Scientists Discover Gene That May Halt Ovarian}}\\
	\textsf{\footnotesize{All Ordinaries Finishes Day On Closing High}}\\
	\textsf{\footnotesize{Waratahs Starting Spot Not A Mere Formality Sailor}}\\
	\textsf{\footnotesize{Proposed Subdivision Wont Affect Recreational}}\\
	\textsf{\footnotesize{Bangladesh To Play Three Tests Five Odis In}}\\
	\textsf{\footnotesize{Minister Promises More Resources To Combat Child}}
\end{center}

Most of these examples contain grammatical errors, such as ending with an adjective, while some headlines contain absurd or nonsensical content. These deficiencies set these headlines apart from the rest. It is worth noting that participants appeared more likely to identify headlines containing grammatical errors as computer-generated than other types of errors.

\section{Classification}\label{sec:classification}

For our classifier experiments, we used the three sets of data (2015, 2016 and 2017) we had previously compiled. Specifically, for training we only used the 2015 set, while the 2016 and 2017 sets were used for testing. Splitting the train and test data by the year of publication ensures that there is no overlap between the sets and there is some variability between the content of the headlines (for example, different topics/authors). Therefore, we can be confident that the classifiers generalize to unknown examples.

Furthermore, for hyperparameter tuning, the 2015 data was randomly split into training and development sets on a $80/20$ ratio. In total, for training there are $129,610$ headlines, for development there are $32,402$ and for testing there are $303,965$.

\subsection{Experiments}

Four types of classifiers were explored: baselines (Elastic Net and Naive Bayes), deep learning (CNN, Bi-LSTM and Bi-LSTM with Attention), transfer learning via ULMFit \cite{ulmfit} and Transformers (BERT \cite{bert} and DistilBERT \cite{distilbert}). The architecture and training details can be found in Appendix \ref{app:class_details}.

Results are shown in Table \ref{table_res}. Overall accuracy is the accuracy in percentage over all headlines (real and generated), while (macro) precision and recall are calculated over the generated headlines. Precision is the percentage of correct classifications out of all the generated classifications, while recall is the percentage of generated headlines the model classified correctly out of all the actual generated headlines. High recall scores indicate that the models are able to identify a generated headline with high accuracy, while low precision scores show that models classify headlines mostly as generated.

We can observe from the results table that humans are overall less effective than all the examined models, including the baselines, scoring the lowest accuracy. They are also the least accurate on generated headlines, achieving the lowest recall. In general, human predictions are almost as bad as random guesses. 

Deep learning models scored consistently higher than the baselines, while transfer learning outperformed all previous models, reaching an overall accuracy of around $83\%$. Transformer architectures though perform the best overall, with accuracy in the $85\%$ region. BERT, the highest-scoring model, scores around $30\%$ higher than humans in all metrics. The difference between the two BERT-based models is minimal.

\begin{table}[!t]
\small
\centering
\begin{tabular}{|
>{\columncolor[HTML]{EFEFEF}}l| c | c | c |}
\hline
\rowcolor[HTML]{b5b5b5}
\multicolumn{1}{|c|}{\cellcolor[HTML]{b5b5b5}Method} & \multicolumn{1}{c|}{\cellcolor[HTML]{b5b5b5}Ovr. Acc.} & \multicolumn{1}{c|}{\cellcolor[HTML]{b5b5b5}Precision} & \multicolumn{1}{c|}{\cellcolor[HTML]{b5b5b5}Recall} \\ \hline\hline
Human   & \textit{50.4} & 66.3  & \textit{52.2} \\ \hline\hline
Naive Bayes & 50.6 & 58.5 & 56.9 \\ \hline
Elastic Net & 73.3 & \textit{58.1} & 62.3 \\ \hline\hline
CNN & 81.7 & 75.3 & 76.2 \\ \hline
BiLSTM & 82.8 & 77.9 & 77.3 \\ \hline
BiLSTM/Att. & 82.5 & 76.9 & 77.2 \\ \hline\hline
ULMFit & 83.3 & 79.1 & 78.5 \\ \hline\hline
BERT & \textbf{85.7} & \textbf{86.9} & \textbf{81.2} \\ \hline
DistilBERT & 85.5 & 86.8 & 81.0 \\ \hline

\end{tabular}
\caption{Each run was executed three times with (macro) results averaged. Standard deviations are omitted for brevity and clarity (they were in all cases less than 0.5).}
\label{table_res}
\end{table}

Since training and testing data are separate (sampled from different years), this indicates that there are some traits in generated text that are not present in human text. Transformers are able to pick up on these traits to make highly-accurate classifications. For example, generated text shows lower variance than human text \cite{gtlr}, which means text without rarer words is more likely to be generated than being written by a human.

\subsection{Error Analysis}

We present the following two computer-generated headlines as indicative examples of those misclassified as real by BERT:

\begin{center}
	\textsf{\footnotesize{Extra Surveillance Announced For WA Coast}}\\
	\textsf{\footnotesize{Violence Restricting Rescue Of Australian}}
\end{center}

The first headline is not only grammatically sound, but also semantically plausible. A specific region is also mentioned (``WA Coast"), which has low probability of occurring and possibly the model does not have representative embeddings for. This seems to be the case in general, with the mention of named entities increasing the chance of fooling the classifier. The task of predicting this headline is then quite challenging. Human evaluation was also low here, with only $19\%$ of participants correctly identifying it.

In the second headline, the word ``restricting" and the phrase ``rescue of" are connected by their appearance in similar contexts. Furthermore, both ``violence" and ``restricting rescue" have negative connotations, so they also match in sentiment. These two facts seem to lead the model in believing the headline is real instead of computer-generated, even though it is quite flimsy both semantically (the mention of violence is too general and is not grounded) and pragmatically (some sort of violence restricting rescue is rare). In contrast, humans had little trouble recognising this as a computer-generated headline; $81\%$ of participants labelled it as fake. This indicates that automated classifiers are still susceptible to reasoning fallacies.

\section{Conclusion}
\label{sec:conclusion}

This paper examined methods to detect headlines generated by a GPT-2 model. A dataset was created using headlines from ABC and a survey conducted asking participants to distinguish between real and generated headlines.

Real headlines were identified as such by $55.9\%$ of the participants, while generated ones were identified with a $47.8\%$ rate. Various models were trained, all of which were better at identifying generated headlines than humans. BERT scored $85.7\%$, an improvement of around $35\%$ over human accuracy.

Our work shows that whereas humans cannot differentiate between real and generated headlines, automatic detectors are much better at the task and therefore do have a place in the information consumption pipeline.

\section*{Acknowledgments} This work was supported by ERCAdG \#740516. We want to thank the anonymous reviewers for their insightful comments and questions, and the members from the University of Sheffield who participated in our survey.

\bibliography{anthology}
\bibliographystyle{acl_natbib}

\appendix

\section{Classifier Details}
\label{app:class_details}

ULMFit and the Transformers require their own special tokenizers, but the rest of the models use the same method, a simple indexing over the most frequent tokens. No pretrained word vectors (for example, GloVe) were used for the Deep Learning models.

ULMFit uses pre-trained weights from the AWD-LSTM model \cite{awd_lstm}. For fine-tuning, we first updated the LSTM weights with a learning rate of $0.01$ for a single epoch. Then, we unfroze all the layers and trained the model with a learning rate of $7.5e$-$5$ for an additional epoch. Finally, we trained the classifier head on its own for one more epoch with a learning rate of $0.05$.

For the Transformers, we loaded pre-trained weights which we fine-tuned for a single epoch with a learning rate of $4e$-$5$. Specifically, the models we used were base-BERT (12 layers, 110m parameters) and DistilBERT (6 layers, 66m parameters).

The CNN has two convolutional layers on top of each other with filter sizes 8 and 4 respectively, and kernel size of 3 for both. Embeddings have 75 dimensions and the model is trained for $5$ epochs.

The LSTM-based models have one recurrent layer with 35 units, while the embeddings have 100. Bidirectionality is used alongside a spatial dropout of 0.33. After the recurrent layer, we concatenate average pooling and max pooling layers. We also experiment with a Bi-LSTM with self-attention \cite{attention}. These models are trained for $5$ epochs.

\end{document}